\documentclass[letterpaper, 10 pt, conference]{ieeeconf}

\IEEEoverridecommandlockouts                            

\usepackage[english]{babel}
\usepackage{amsmath} 
\usepackage{amssymb}  
\usepackage{graphicx}
\usepackage{subfigure}
\usepackage{caption}
\usepackage{multirow}
\usepackage{amsfonts}
\usepackage{hyperref}
\usepackage{tabularx}
\usepackage[noend]{algpseudocode}
\usepackage{algorithm}
\usepackage{bm}
\usepackage{breakurl}
\usepackage{enumerate}
\usepackage{verbatim}
\usepackage{float}
\usepackage{siunitx}
\usepackage{mdframed}
\usepackage{adjustbox}
\usepackage{cleveref}
\usepackage{authblk}
\usepackage{color}
\usepackage{soul}

\usepackage{booktabs}  

\title{\LARGE \bf
3D Reconstruction-Based Seed Counting of Sorghum Panicles \\
for Agricultural Inspection
}

\author{
Harry Freeman$^{1*}$, Eric Schneider$^{1*}$, Chung Hee Kim$^{1}$, Moonyoung Lee$^{1}$, George Kantor$^{1}$
\thanks{$^{1}$Carnegie Mellon University Robotics Institute, PA, USA 
        \texttt{\{hfreeman, franzs, chunghek, moonyoul, kantor\}@cs.cmu.edu}}

\thanks{$^{*}$These authors contributed equally to this work.}
}

\begin{document}
\maketitle
\thispagestyle{empty}
\pagestyle{empty}



\begin{abstract}
In this paper, we present a method for creating high-quality 3D models of sorghum panicles for phenotyping in breeding experiments. This is achieved with a novel reconstruction approach that uses seeds as semantic landmarks in both 2D and 3D. To evaluate the performance, we develop a new metric for assessing the quality of reconstructed point clouds without having a ground-truth point cloud. Finally, a counting method is presented where the density of seed centers in the 3D model allows 2D counts from multiple views to be effectively combined into a whole-panicle count. We demonstrate that using this method to estimate seed count and weight for sorghum outperforms count extrapolation from 2D images, an approach used in most state of the art methods for seeds and grains of comparable size.
\end{abstract}


\section{Introduction}

With recent advancements in data-driven computer vision, agriculture is widely adopting image-based techniques to efficiently inspect vast quantities of crops. Automated crop inspections, which were not easily done before, enable farmers and breeders to make real-time decisions to manage pests, disease, and drought, and to automate laborious tasks such as phenotyping and yield prediction. 
In this paper we propose a computer vision-based method for non-destructive counting of sorghum seeds for early forecasting of yield. Accurate forecasting is valuable for sorghum breeding programs, as it would allow faster decision-making on variant suitability, which could expedite the current five-year breeding process \cite{house1985guide}. Seed count would be a valuable phenotypic trait, but it is currently not possible to sample in a non-destructive way.

Common deep learning techniques for visual agricultural inspection include disease classification \cite{disease}, object detection \cite{cropdeep}, and fruit counting \cite{sorghum2020}.
In contrast to the large and separated fruits typically inspected, we investigate seed modelling on a sorghum panicle, which is more challenging from a computer vision perspective. The seeds are much smaller than typically studied crops (average diameter 3.3mm), making them difficult to detect and track. In addition, there is significantly more occlusion due to dense packing and clutter from husks. Although there has been work on 2D image based instance counts for other crops \cite{pmid34198797, 8716704, khaki2021deepcorn}, it is still difficult to obtain a high accuracy count with sorghum.
We create a high-quality 3D model from stereo using a semantic landmark-based reconstruction approach, which we use to count seeds. Using our proposed method, we acquire a more realistic count than 2D image-based approximations.

\vspace{7pt}
The specific contributions of this paper are:
\begin{itemize}
    \item A novel 3D reconstruction method that utilizes seeds as semantic landmarks in 2D and 3D to produce a high quality model of a sorghum panicle. 
    \item A new metric for assessing point cloud reconstruction quality in the absence of ground truth.
    \item A novel method for extracting seed counts from point clouds by extending 2D image processing techniques into 3D, along with an algorithm to identify local maxima in a point cloud.
    \item A dataset of sorghum stereo images with camera poses, a subset labeled with instance segmentation of seeds. \footnote{\url{https://labs.ri.cmu.edu/aiira/resources/}}.
\end{itemize}

\begin{figure*}[!ht]
    \centering
    \includegraphics[width=\linewidth]{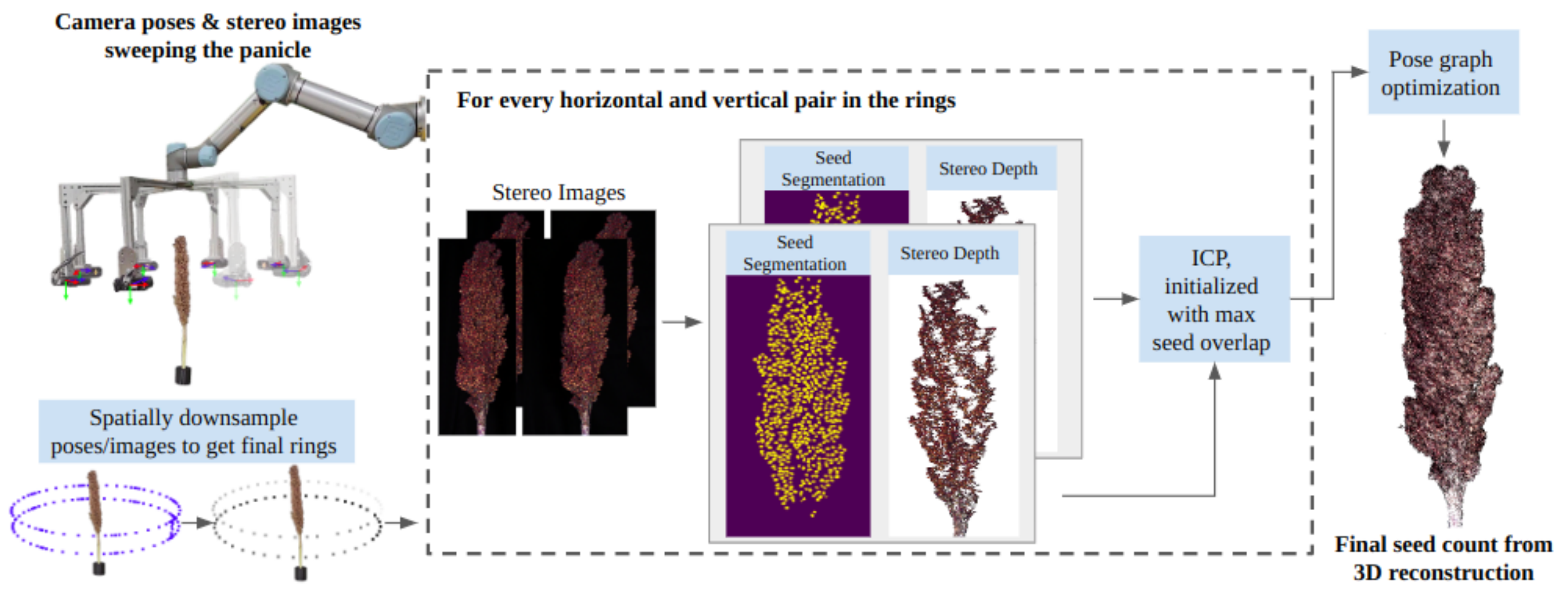}
    \caption{3D Reconstruction pipeline for the sorghum stalk}
    \label{fig:reconstruction_pipeline}
    \vspace{-10pt}
\end{figure*}

\section{Related Work}

There has been a significant amount of recent work dedicated towards reconstruction and counting in agricultural settings. Mapping and estimating the yield of mangoes in occluded environments using a FasterRCNN segmenter is presented in \cite{stein2016image} and \cite{liu2019monocular}. Mapping and counting grapes in 3D by fitting spheres to point clouds
is presented in \cite{nellithimaru2019rols}. While these methods work in their respective domains, they do not extend well to sorghum where the seeds are smaller and the level of density and occlusions are higher, making them harder to consistently segment and fit shapes to.

Phenotyping during the breeding process is laborious if done manually and important for expedited decision making. As a result, several works address automated phenotyping using robots. In \cite{varela_sorghum_UAV_2021}, UAV images taken early in the season are used to predict end of season above-ground biomass. \cite{bao_sorghumpheno_2019} and \cite{Young2019} show that images collected from mobile robots can be used to assess plant height and stalk size more easily than manual collection. Component traits such as these are used in genetic research to improve biomass yield. Our work explores seed counting, which was not possible at the resolution of these systems.

There has also been relevant work in estimating seed counts for smaller crops from single 2D images.  
Counting rice and soybeans with density maps using convolutional neural networks is addressed in \cite{pmid34198797} and \cite{8716704} respectively. However, the rice and beans have been stripped from the plant and laid out such that there are few occlusions.
Density maps have also been used to count corn kernels on the cob, where the final count is proportional to the density map count as a result of corn's symmetric shape \cite{khaki2021deepcorn}. Similarly, \cite{Nuske2014AutomatedVY} uses a KD-Forest approach to detect grapes in clusters using keypoint-based features, and estimates yield using a scale factor. These methods do not adapt well to sorghum seeds because of the asymmetric nature of sorghum panicles.

With regards to reconstruction in agriculture, most works are focused on larger maps and fields rather than single plants. For example, large field maps of different crops are reconstructed in \cite{potena_2019_agmapping} and \cite{chebrolu_2017_agmapping}. Although localized views of flowers and vines are captured in \cite{ohi_2018_agmapping} and \cite{silwal2021bumblebee}, they do not get a complete 360$^\circ$ scan. Apple orchard rows are reconstructed in \cite{roy2018registering} by merging views from opposing sides using cylinders fit to trunks. This does not adapt well to sorghum as the stems are too small to effectively fit.


\section{Methodology}~\label{sec:methodology}

\vspace{-15pt}

\subsection{Overview}\label{sec:overview}

In order to generate a high-quality 3D model of sorghum panicles,
we set up an automatic data collection process by attaching a flash stereo camera \cite{9636542} to the wrist of a UR5 robot arm. The robot follows a circular trajectory around the panicle as shown in Fig.~\ref{fig:reconstruction_pipeline}. Although these images were captured in the lab, in-field image capture from an arm mounted on a mobile base would also be possible. From all images taken, we spatially downsample to only consider images $\mathbf{I}_i \in \mathbb{I}$ and poses $\mathbf{T}_i \in \mathbb{T}$ in the shape of a double ring, spaced 5cm apart, as seen in Fig.~\ref{fig:reconstruction_pipeline}, leaving roughly 85 images per panicle. We then use RAFT-Stereo \cite{lipson2021raft} to construct point clouds for each frame. Using Iterative Closest Point (ICP) on segmented seeds alone, we construct a pose graph that best aligns all point clouds to create the final high quality point cloud $\mathbf{C}$. Lastly, given $\mathbf{C}$, seed masks are combined between all images $\mathbf{I}_i$ to obtain a final seed count.

\subsection{Instance Segmentation}\label{sec:detection}

Given a stereo image pair, we acquire a 3D point cloud semantically labeled with individual sorghum seeds. This is achieved through instance segmentation on 2D images, which is projected onto the 3D points.
Our seed segmentation is based on a CenterMask \cite{lee2020centermask} instance segmentation network. 
Seeds were hand segmented from 10 different $1440\times1080$ sorghum images across different species. After training, seed masks are projected onto the 3D point cloud.

\subsection{Global Registration}\label{sec:ICP}

\begin{figure}[!thbp]
    \vspace{-2pt}
    \centering
    \includegraphics[width=0.88\linewidth]{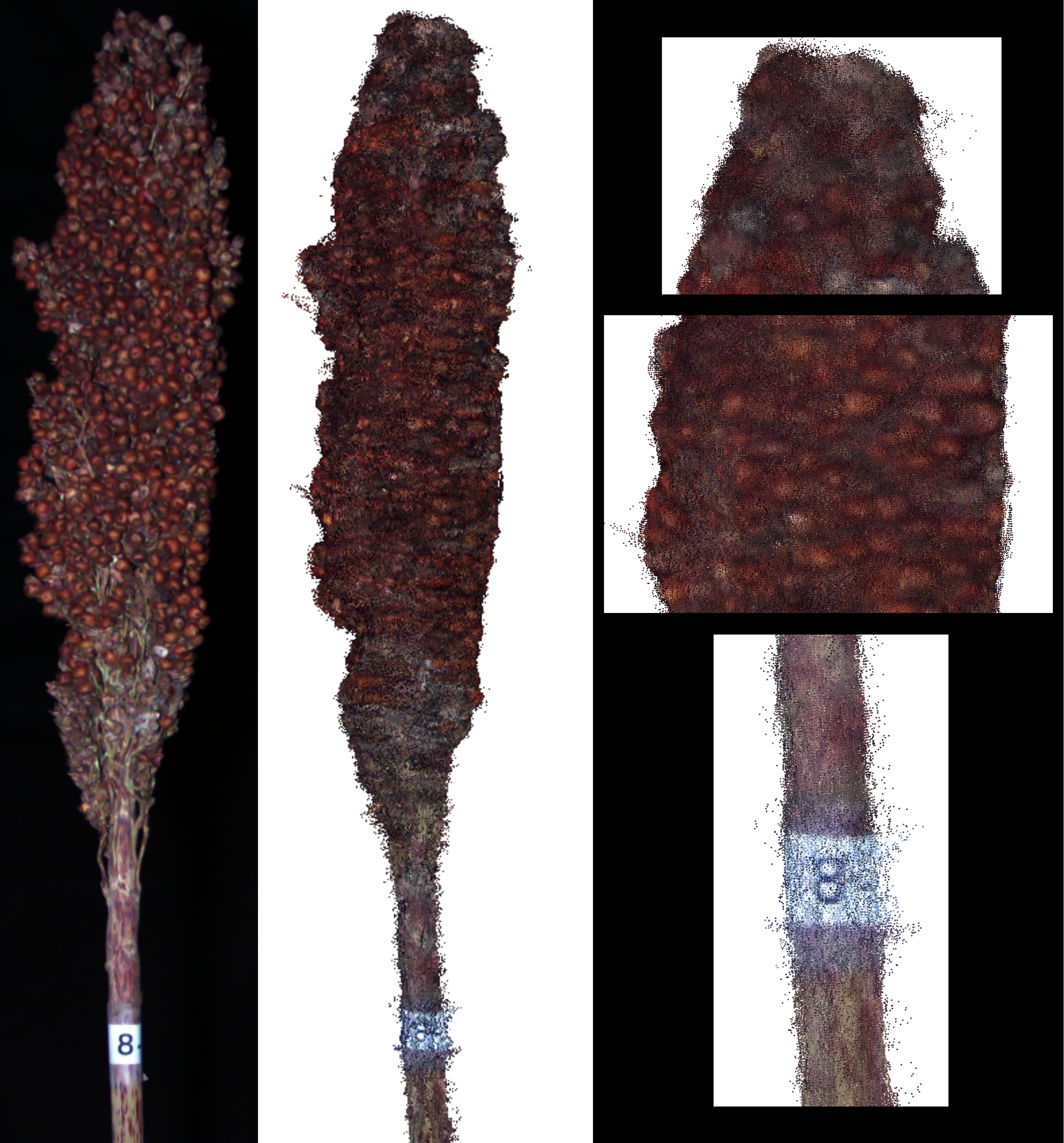}
    \put(-214,5){\textcolor{white}{(a)}}
    \put(-162,5){(b)}
    \put(-92,5){\textcolor{white}{(c)}}
    \caption{Example reconstruction results. (a) one of the original RGB images, (b) the colorized point cloud, (c) zoomed view of the colorized point cloud at the stem, mid-body, and tip. Some points of interest include the ``8" on the stem label, and the body outline which matches the RGB outline well.}
    \label{fig:full_reconstruction}
\end{figure}

We jointly register point clouds of a sorghum panicle imaged from different viewpoints via pose graph optimization, as presented in \cite{7299195}. One challenge is that the point clouds are dense, and ICP optimization on the full cloud performed poorly due to bad correspondences, an example of ICP falling into local minima. Instead we choose a limited set of high-quality points in the cloud and run ICP only on those points, somewhat analogous to doing optical flow on higher quality landmarks like SIFT features. Semantically segmented seeds with high confidence are identified based on their inference scores. The set of good seeds from image $\mathbf{I}_i$ are then used as node $\mathbf{P}_i$ in the pose graph, and pose graph optimization is performed using the Levenberg-Marquardt algorithm \cite{levenberg1944method}. An example of a reconstructed panicle is shown in Fig.~\ref{fig:full_reconstruction}(b).

\begin{figure}[!thbp]
    \centering
    \includegraphics[width=\linewidth]{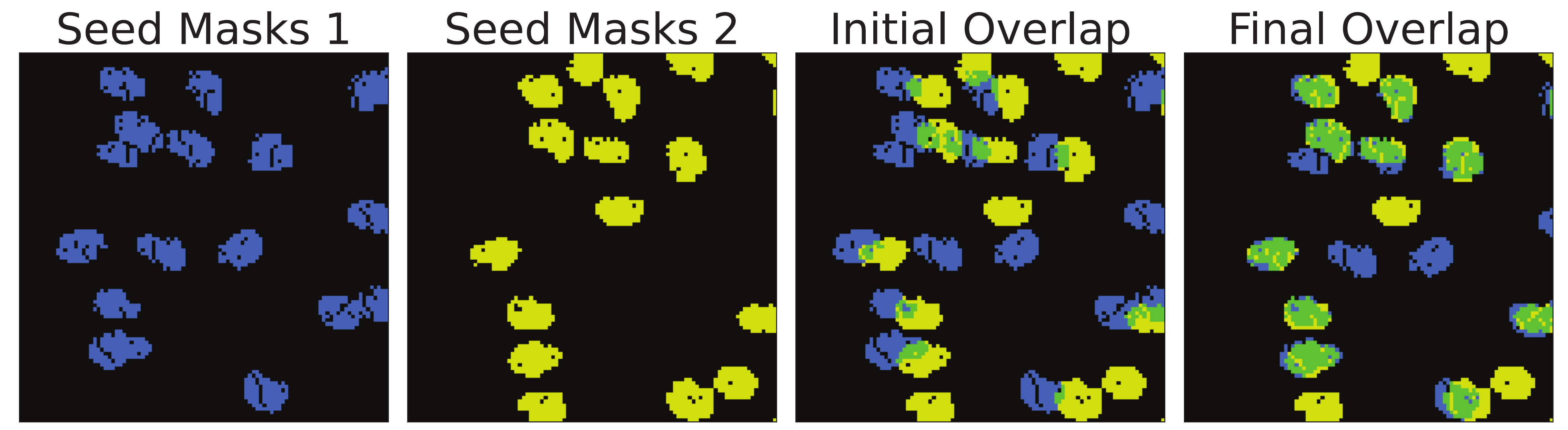}
    \caption{Matching mask structure with maximum IOU. Seed masks 1, seed masks 2, and their intersection are colored blue, yellow, and green in respective order.}
    \label{fig:IOU_process}
    \vspace{-10pt}
\end{figure}

We observe that using camera poses from arm kinematics to initialize ICP yields poor results on the scale of seeds. This is due to error in extrinsic camera parameters, despite using a standard hand-eye calibration process.
Hence, we refine the camera pose priors by maximizing seed mask overlap. The seed masks of two neighboring nodes $\mathbf{P}_i$ and $\mathbf{P}_j$ are projected into a common image frame, at the average pose between $\mathbf{T}_i$ and $\mathbf{T}_j$. We search for the pixel shifts that yield maximum intersection over union (IOU) of seed masks as shown in Fig.~\ref{fig:IOU_process}. The \textit{No Shift Maximize} ablation test in Fig.~\ref{fig:assess_3d} shows that this IOU maximization improves reconstruction.

\subsection{Counting}\label{sec:counting}

In order to obtain a final seed count, we use the 3D model to ensure that a single true seed segmented in multiple images will be counted only once. The following new 3D counting method is proposed to perform this combination of 2D counts while handling the close proximity of neighboring seeds, spurious detections, and noise in the point cloud.

\begin{figure}[!thbp]
    \centering
    \includegraphics[width=0.9\linewidth]{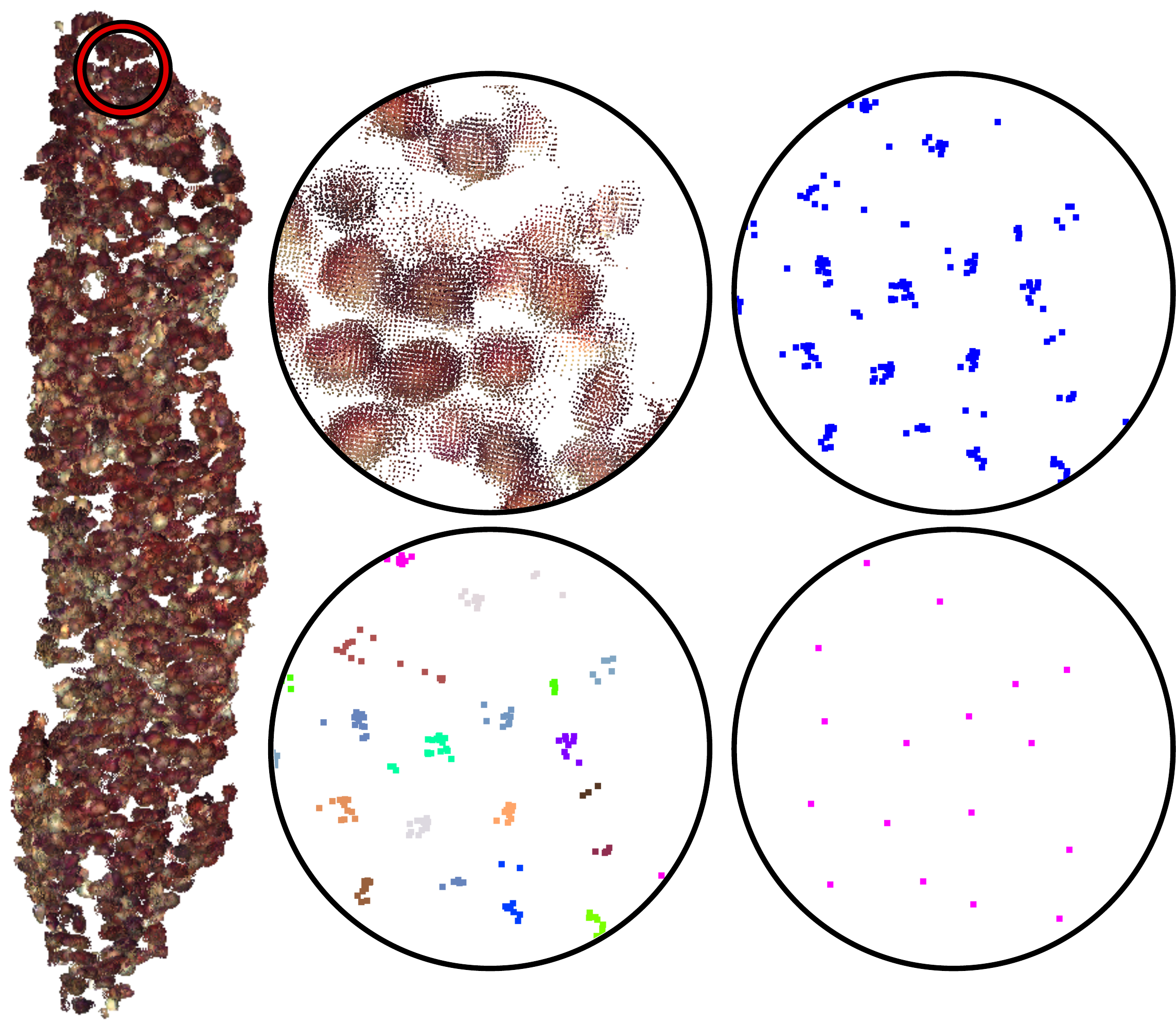}
    \put(-185,5){(a)}
    \put(-99,97){(b)}
    \put(-5,100){(c)}
    \put(-102,5){(d)}
    \put(-5,5){(e)}
    \caption{(a) An example of a final point cloud seed mask, (b) zoomed seeds, (c) seed centers, (d) seed centers clustered with DBSCAN, and (e) final seed sites.}
    \label{fig:seed_mask_seed_dbscan}
\end{figure}
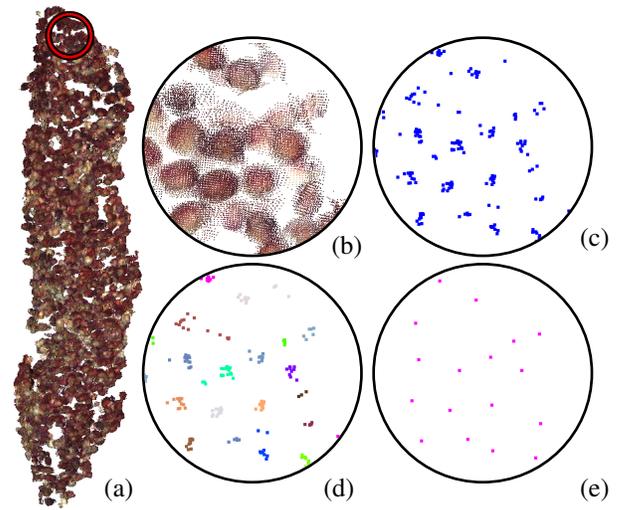
\begin{figure}[!thbp]
    \centering
    \includegraphics[width=0.9\linewidth]{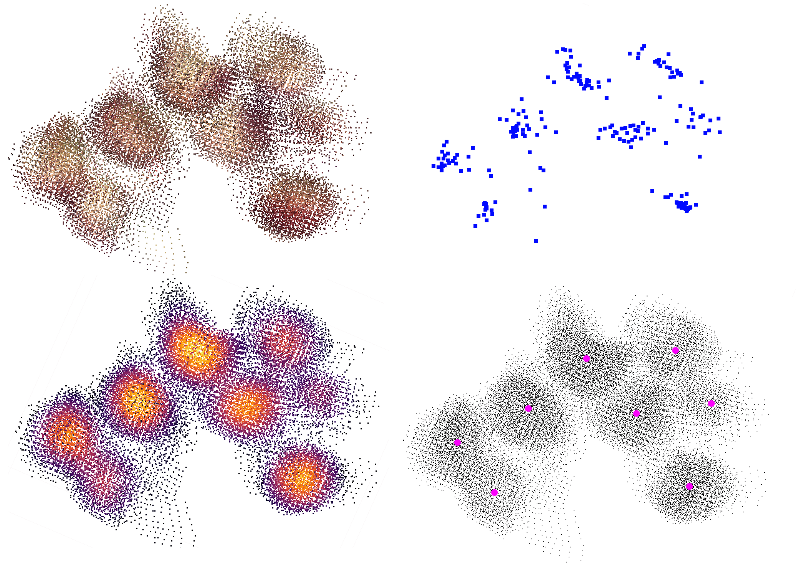}
    \put(-170,100){(a)}
    \put(-65,100){(b)}
    \put(-165,15){(c)}
    \put(-65,15){(d)}
    \caption{(a) Seed point cloud that has been put in a single cluster by DBSCAN, (b) seed centers from individual images, (c) seed points weighted by seed-center density, and (d) local maxima (pink) that have been chosen as seeds.}
    \label{fig:mult_seed_example}
    \vspace{-10pt}
\end{figure}

First, 3D seed centers are clustered using density-based spatial clustering (DBSCAN) \cite{ester1996densitybased}, as shown in Fig.~\ref{fig:seed_mask_seed_dbscan}(d). Seeds are then counted in each cluster, which breaks up the large problem of counting thousands of seeds across the whole panicle as there are often fewer than four true seeds in a given cluster.
Next, we adapt the concept of 2D image smoothing and apply it to 3D point clouds. In image processing, a 2D Gaussian filter smooths an image by calculating a weighted average around each pixel's neighborhood.
In our method, each seed center in the cluster is treated as a unit-impulse, and each impulse is smoothed around a volume of space using a 3D Gaussian filter.
An example of this density map can be seen in Fig.~\ref{fig:mult_seed_example}(c).

Once the density values of the mask cloud points are created, the final step to calculate the number of seeds in each cluster is to find the local maxima within a defined radius. This is a type of non-maximal suppression (NMS) on the density values.
Each local maximum corresponds to a unique seed and is the location of the seed's center. Algorithm \ref{alg:local_max} has a detailed description of the proposed algorithm.

\algrenewcommand\algorithmicrequire{\textbf{Inputs:}}
\algrenewcommand\algorithmicensure{\textbf{Output:}}

\begin{figure}[!thbp]
\vspace{-10pt}
\begin{minipage}{0.48\textwidth}
\renewcommand*\footnoterule{}
\begin{algorithm}[H]
\caption{Find Cluster Local Maxima}
\label{alg:local_max}
\let\thefootnote\relax\footnotetext{Summary: All points $p$ in the cluster's cloud $M$ are initialized as unvisited (line 1). The unvisited point with the highest density value $D(p)$ is iteratively retrieved (line 4), and if that point has a higher density value than all neighbors in a defined radius $r$, it is a local maximum (lines 6-9). All points within the radius are marked as visited (line 10), and the process repeats until all points have been visited.}
\begin{algorithmic}[1]
\Require $M \in \mathbb{R}^{k \times 3}$, $r \in \mathbb{R}^{1}$, $D$
\Ensure $L \in \mathbb{R}^{l \times 3}$

\State $U \gets M, L \gets \emptyset$

\While{$U \neq \emptyset$} 
    \State $s \gets \max_{p \in U} D(p)$
    \State $U \gets U \setminus \{s\}$
    \State $R \gets \{p \mid p \in M \text{ and } 0 < \| s - p\| < r\}$
    \State $m \gets \max_{p \in R} D(p)$
    \If{$D(s) > D(m)$}
        \State $L \gets L \cup \{s\}$
        
    \EndIf
    
    \State $U \gets U \setminus \{p \mid p \in R\}$
\EndWhile
\end{algorithmic}
\end{algorithm}
\end{minipage}
\end{figure}


Once all local maxima are found for each cluster, the total number of maxima becomes the final seed count. An example of this process on a single cluster can be seen in Fig.~\ref{fig:mult_seed_example}.

\subsection{3D Reconstruction Metric}\label{sec:recon_metric}

    Several prior works \cite{zhao_2012_shading}, \cite{zhang_2021_reconstruction} discuss quantitative reconstruction evaluation in the absence of ground truth, but they require that the final output to evaluate against is a mesh. Our reconstruction method produces a dense point cloud, so we developed and validated a novel cloud-only rendering based method for assessing reconstruction quality in the absence of ground truth. We compare a small circle of pixels sampled from an RGB image $\mathbf{I}_i \in \mathbb{I}$, centered on a sampled seed, against a projected render of the same seed made using the full reconstructed cloud. A sampling function $\lambda$ is defined so that $K$ seeds are sampled per image along the center of the vertical axis where the projections are cleanest. This method experimentally indicates relative levels of noise in the reconstructed point clouds by comparing rendered sections to the original RGB images.

\begin{figure}[!thbp]
        \vspace{-5pt}
        \centering
        \includegraphics[width=0.9\linewidth]{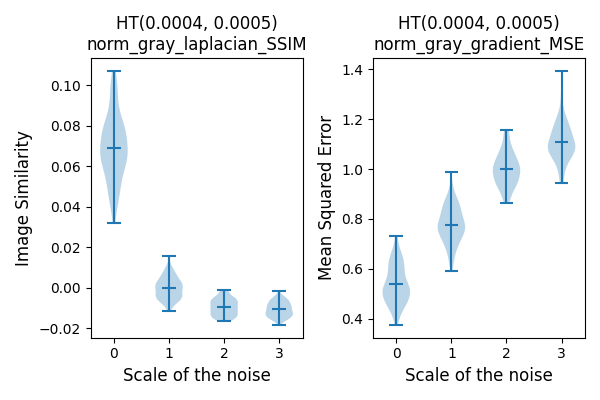}
        \caption{Response of chosen metrics to introduced noise. Noise took the form of homogeneous transforms, with $(x, y, z)$ translations drawn from a Gaussian $\mathcal{N}(0, \sigma=\text{scale}*0.4\text{mm})$ and rotational noise $(\theta, \phi, \psi)$ drawn from a Gaussian $\mathcal{N}(0, \sigma=\text{scale}*0.0005\text{rad})$. After the random transforms were applied the cloud was recalculated and rendered.}
        \label{fig:3d_metric}
    \end{figure}
\begin{figure}[!thbp]
        \centering
        \includegraphics[width=0.65\linewidth]{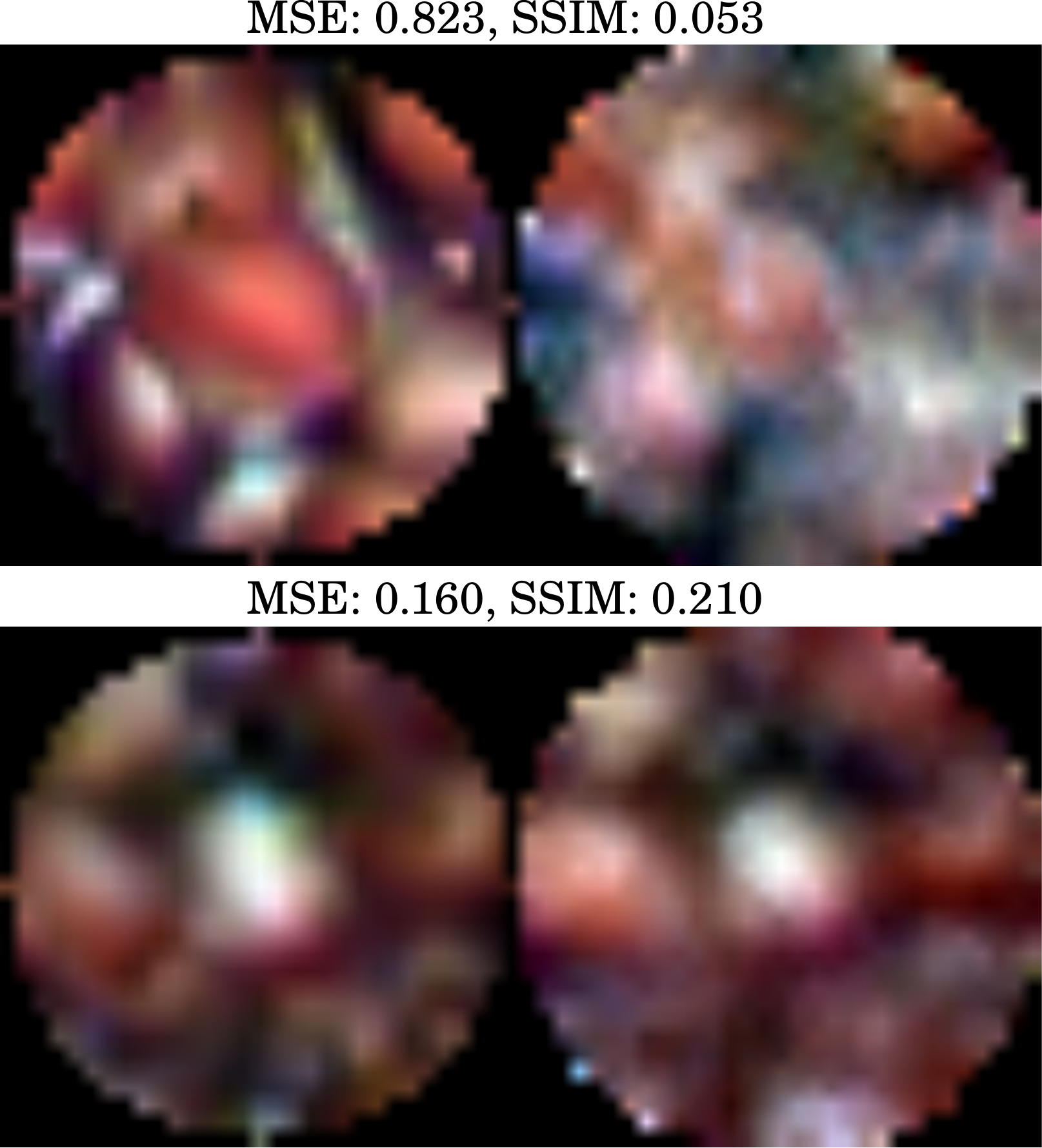}
        \caption{Qualitative examples of the reconstruction metrics, drawn from low and high scoring samples. On the left are image patches from the original RGB images, on the right are image patches rendered from the reconstructed point cloud. All patches are normalized so each channel has min/max values of 0/255.}
        \label{fig:3d_metric_examples}
        \vspace{-35pt}
    \end{figure}

    To validate this framework, noise was purposefully introduced in the camera poses $\mathbf{T}_i$ when creating the reconstructed cloud. A variety of comparisons were run on pairs of RGB image patches and the corresponding rendered patches. The strongest response to introduced noise came from normalized grayscale image patches. Both the mean-squared error (MSE) on image gradients, and the Structural Similarity \cite{wang_2004_SSIM} (SSIM) on image Laplacians responded well to the introduced noise, shown in Fig.~\ref{fig:3d_metric}. In order to settle on these two metrics we checked all combinations of RGB/grayscale, normalized/unnormalized, and intensity/gradient/Laplacian. Two examples of our image-to-render comparison with their corresponding MSE and SSIM scores are shown in Fig.~\ref{fig:3d_metric_examples}.

    Our reconstruction quality metrics ``$\alpha\beta$-MSE" and ``$\alpha\beta$-SSIM" are defined as follows. For each image, $\lambda$ samples $K$ seeds from $\mathbb{S}_i$, where $\mathbb{S}_i$ are the seeds in image $\mathbf{I}_i$. For a sampled seed $s_{ik} \in \mathbb{S}_i$, the image patch $\alpha_{ik}$ and rendering of the point cloud $\beta_{ik}$ are generated, both of which are grayscaled and normalized. The MSE and SSIM of $\alpha_{ik}$ and $\beta_{ik}$ are calculated, then averaged over all seeds and panicles.
    \vspace{-10pt}

    \begin{align*}
        \text{MSE}_{ik} &= \frac{1}{N} \sum_{pixels} \big[ \nabla \alpha_{ik} - \nabla \beta_{ik} \big]^2\\
        \alpha\beta \text{-MSE} &= \frac{1}{P} \sum_p \frac{1}{IK} \sum_i \sum_{k \in \lambda(\mathbb{S}_i)} \text{MSE}_{ik} \\
        \text{SSIM}_{ik} &= \text{SSIM} \big( \mathcal{L} \big(\alpha_{ik}\big), \mathcal{L} \big(\beta_{ik}\big) \big)\\
        \alpha\beta \text{-SSIM} &= \frac{1}{P} \sum_p \frac{1}{IK} \sum_i \sum_{k \in \lambda(\mathbb{S}_i)} \text{SSIM}_{ik}
    \end{align*}

    Here $\nabla$ is the image gradient, $\mathcal{L}$ is the image Laplacian, $\frac{1}{IK} \sum_i \sum_{k \in \lambda(\mathbb{S}_i)}$ indicates an average over sampled seeds in all images, and $\frac{1}{P} \sum_p$ indicates an average over all panicles.
\section{Experimental Results}~\label{sec:experiments} 

\vspace{-20pt}
\subsection{Dataset}\label{sec:dataset}

\begin{figure}[!t]
    \centering
    \includegraphics[width=0.9\linewidth]{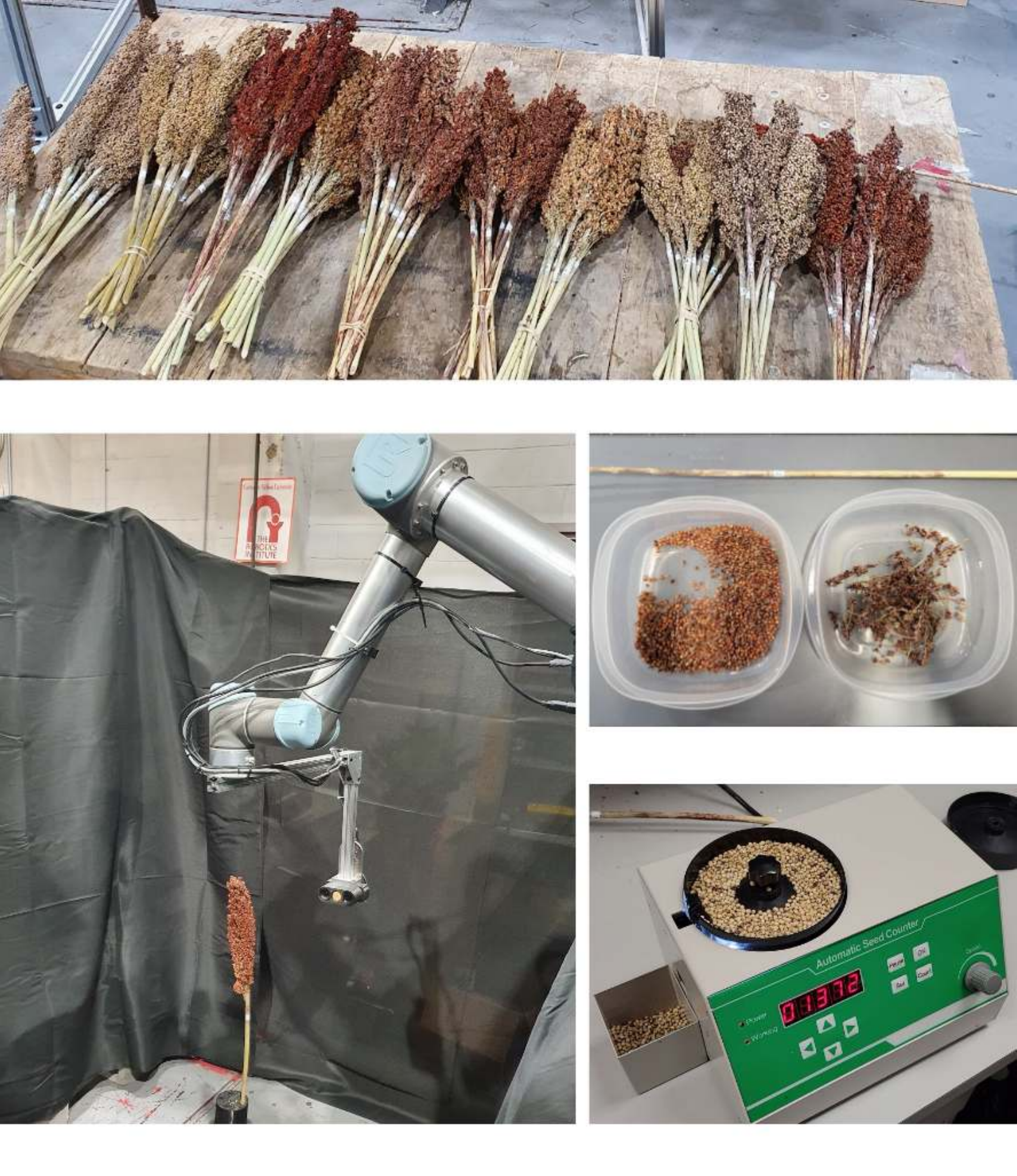}
    \put(-115,165){\small (a)}
    \put(-160,0){\small (b)}
    \put(-52,89){\small (c)}
    \put(-52,0){\small (d)}
    \caption{(a) 100 sorghum panicles from 10 different sorghum species. (b) Our data collection system, a stereo camera attached to the UR5 robot arm. (c) Seeds were manually stripped and (d) counted using a seed counting machine.}
    \label{fig:data_collection}
    \vspace{-10pt}
\end{figure}

Our dataset consists of stereo images of 100 sorghum panicles collected in two $360^\circ$ rings using a custom stereo camera \cite{9636542}. There were 10 panicles from 10 different species as seen in Fig.~\ref{fig:data_collection}(a).
To evaluate our proposed method, we manually stripped panicles (Fig.~\ref{fig:data_collection}(c)) and counted all seeds using an automatic seed counting machine\footnote{Wadoy Automatic Seeds Counter, Sly-C} (Fig.~\ref{fig:data_collection}(d)), which serves as ground truth. The process of stripping seeds, removing husks, and counting took significant effort, on average 40 minutes per panicle, which reinforces the usefulness of an automated method for yield estimation.
Random errors in the seed count include some lost seeds that fell off panicles between image collection and hand-counting. Affecting the count in the opposite direction, some unremoved husks were counted as seeds by the counting machine despite manual efforts to separate seeds from husks. We expect the effect on the ground truth to be small. The stereo images, camera poses, human-labeled seed segmentations, panicle weights, and human-counted seed counts can be found in our dataset\footnote{\url{https://labs.ri.cmu.edu/aiira/resources/}}.

\subsection{3D Reconstruction Quality}\label{sec:assess_3d}
\begin{figure}[!thbp]
        \centering
        \includegraphics[width=0.95\linewidth]{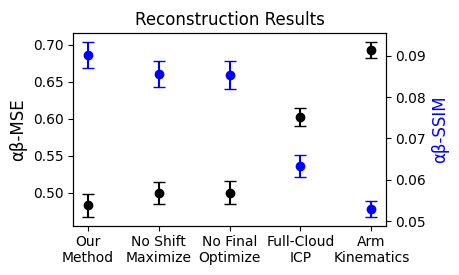}
        \caption{Noise metric results showing growing error and dropping similarity for reconstruction experiments. The vertical bars are the 95\% confidence intervals for the mean of the per-panicle scores.}
        \label{fig:assess_3d}
        \vspace{-10pt}
    \end{figure}

    We assess the effectiveness of our approach with ablation tests using the reconstruction metrics described in \ref{sec:recon_metric}. Below references to ``$\alpha\beta$-MSE" and ``$\alpha\beta$-SSIM" are referring to these specific operations on image and rendered patches. Note that growing $\alpha\beta$-MSE (error) and dropping $\alpha\beta$-SSIM (similarity) both indicate a worse match. Fig.~\ref{fig:assess_3d} shows results of ablation and comparison tests on reconstruction quality.

    \begin{enumerate}
        \item \textit{Our Method}: Our final method, as described in Section \ref{sec:methodology}. All experiments below are tweaks to this approach. Averaged across all panicles, this had the best $\alpha\beta$-MSE and $\alpha\beta$-SSIM scores.


        \item \textit{No Shift Maximize}: The mask overlap maximization discussed in Section \ref{sec:ICP} is not used. This resulted in a slight decrease in reconstruction quality.

        \item \textit{No Final Optimize}: The pair-wise ICP transformations discussed in Section \ref{sec:ICP} are still used to adjust cameras relative to the first frame, but the final optimization is not applied.
        

        \item \textit{Full-Cloud ICP}: Instead of running pair-wise ICP on masked seed points, ICP as described in Section \ref{sec:ICP} was run on the full point clouds. This test showed a significant drop in reconstruction quality.


        \item \textit{Arm Kinematics}: Views were combined using the arm kinematics, with no pose optimization. Although kinematically reconstructed panicles could be used for applications like collision avoidance, they had the worst reconstruction scores and could not be used for counting. Single seeds were clearly represented in multiple 3D locations, ``smeared" cylindrically around the panicle.
    \end{enumerate}

    The best reconstruction results came from pose adjustment using ICP on points determined to be high-quality seeds, and did notably better than ICP naively done using the full cloud from each image. Our hypothesis on why full-cloud ICP is worse is that sorghum is very organic and complex, and picking out meaningful, high-quality areas for ICP to operate on reduces the likelihood of ICP falling into a local minimum. As was discussed in \textit{Arm Kinematics}, the required quality of reconstruction depends on your application. When using 3D structure to identify overlaps in 2D segmentation, decreasing reconstruction quality will lead to counting errors as identifications of the same seed drift apart in space.

\subsection{Prediction of Sorghum Characteristics}\label{sec:assess_count}

\begin{figure}[!thbp]
        \centering
        \includegraphics[width=0.75\linewidth]{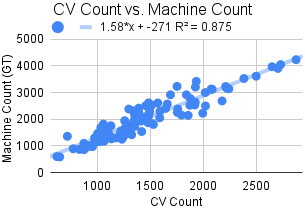}
        \caption{Fit between our method's count (CV Count) and the ground truth count as described in Section \ref{sec:dataset}.}
        \label{fig:count_fit}
    \end{figure}

    As shown in Fig.~\ref{fig:count_fit}, the count produced by our method has a strong linear fit to the ground truth count, with an $R^2$ of 0.875. The 10-fold RMSE using a 75/25 train/test split calculates an average prediction error of 295 seeds. 
    There will always be some error in non-destructive counts, since sorghum panicles have internal, hidden seeds that cannot be seen from an outside view. The only way to expose all seeds is to strip them off the panicles, a time-consuming process.

\begin{figure}[!thbp]
        \centering
        \includegraphics[width=0.75\linewidth]{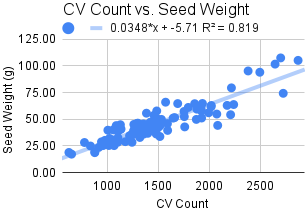}
        \caption{Fit between counted seeds and seed weight, which is the weight of seeds after they have been stripped off a panicle and cleaned of husks.}
        \label{fig:weight_fit}
        \vspace{-10pt}
    \end{figure}

    Ultimately, the characteristic most worth measuring for sorghum is its yield weight, which represents a sellable quantity of the crop. The fit between count and seed weight is still reasonably representative, with an $R^2$ linear fit of 0.819 in Fig.~\ref{fig:weight_fit}, but it fits slightly less well than Fig.~\ref{fig:count_fit}. The 10-fold RMSE using a 75/25 train/test split calculates an average prediction error of 8.5 grams per panicle. This may be due to variations in seed weight. 

\subsection{Benefits of 3D Data over 2D}\label{sec:value_add_3d}

\begin{figure}[!thbp]
        \centering
        \includegraphics[width=\linewidth]{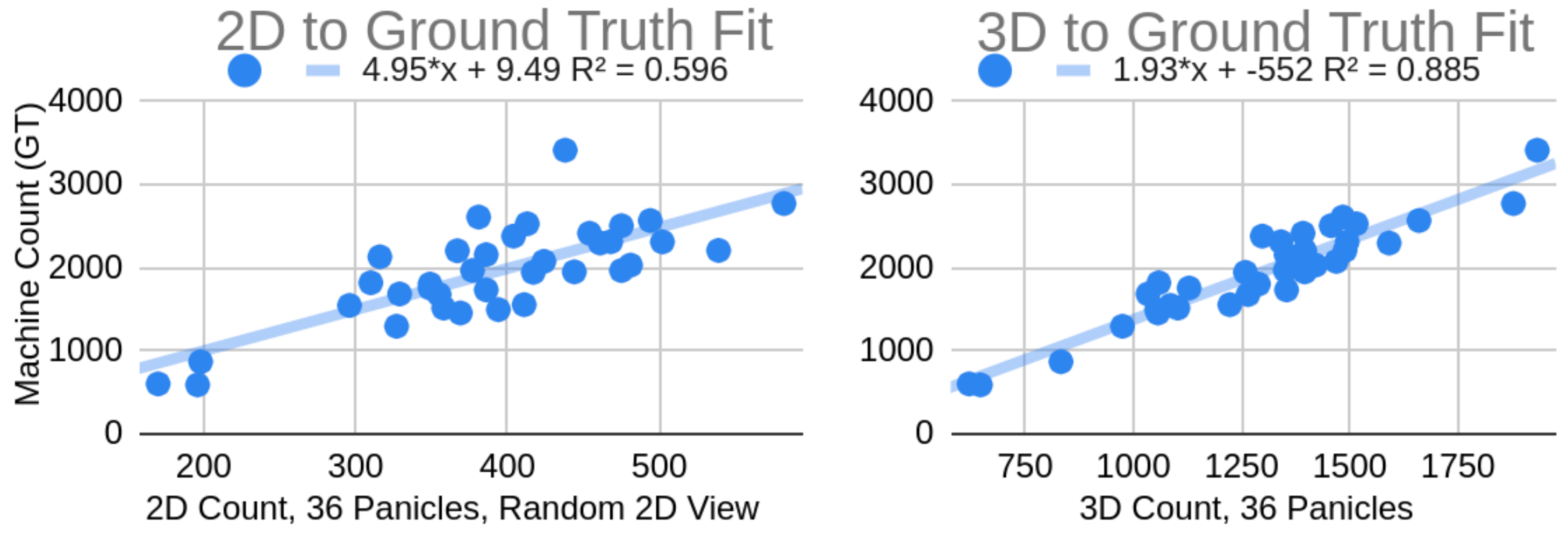}
        \caption{Comparison of 2D and 3D counts fit to ground truth. 2D count comes from a single image per available panicle and has a lower $R^2$ score, indicating worse predictive performance for linear regression. The 10-fold RMSE for these 2D and 3D counts are 353 and 204 respectively.}
        \label{fig:2d3d_linear_fits}
        \vspace{-10pt}
    \end{figure}

\begin{figure}[!thbp]
        \centering
        \includegraphics[width=0.85\linewidth]{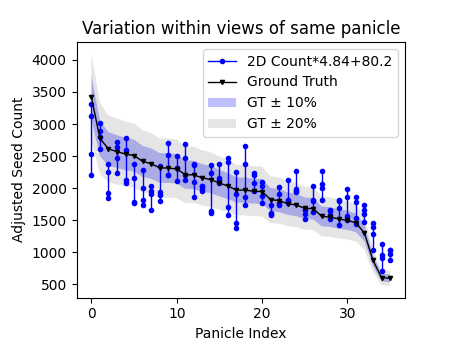}
        \caption{Variation across viewpoints among the 36 panicles, using a linear fit to extrapolate from 2D count to an estimated full count. Linear fit parameters have been recalculated to use four 90$^\circ$ separated images per panicle instead of a random one as in Fig.~\ref{fig:2d3d_linear_fits}. $R^2$ on the increased views was 0.634.}
        \label{fig:2d3d_variation}
        \vspace{-10pt}
    \end{figure}

    In \cite{khaki2021deepcorn} it was shown that it is sufficient to take a 2D count of one side of an ear of corn and scale that to a full kernel count. To test this, ears were rotated around their long axis by 90$^\circ$ increments and imaged, and it was found that single-image kernel counts had low variation because kernels were generally evenly distributed. In contrast, sorghum is more complex in shape, and therefore has more variation when a full count is extrapolated from a single image. In Fig.~\ref{fig:2d3d_linear_fits} and Fig.~\ref{fig:2d3d_variation} we compare the predictiveness of 2D and 3D counts. It may seem unfair to compare 2D and 3D extrapolation because 3D methods have more data available (dozens of images vs. a single image), but it is important to evaluate for hardware considerations. Getting images surrounding a plant for 3D reconstruction is more costly in terms of system complexity, requiring the camera to be actuated rather than fixed to a mobile base such as a tractor, so it is important to assess what relative benefit the 3D method brings.

    In order to test the extrapolation principle, we obtained 2D segment counts from images spaced 90$^\circ$ apart. This was complicated by the fact that some panicles were too tall to be captured in a single frame. To avoid trying to combine segmentation counts from multiple images, we only use counts where the full panicle is visible in four 2D views. 36 out of the 100 panicles met this criteria, enough to get a reasonable representation.

    As seen in Fig.~\ref{fig:2d3d_linear_fits}, 3D counts have a significantly better linear fit to the ground truth counts, with an $R^2$ of 0.885 compared to 0.623 for 2D counts (sampled randomly from the 90$^\circ$ separated views), demonstrating that 3D count is a better predictor of the desired feature.
    The variation in 2D count within each panicle can be seen in Fig.~\ref{fig:2d3d_variation}. There are significant variations in extrapolated counts within each panicle, often stretching to 20-40\% of the ground truth value.

\section{Conclusion}~\label{sec:conclusion} 


One of the benefits to this approach is the integration of segmentation counts across multiple 2D views, using the 3D model to determine which detections are unique. Future detection and segmentation improvements could be folded in to improve estimates while still getting the benefit of view combination. However, the use of multiple views is an intensive process, and uses many images of each panicle. It would be worthwhile to find the minimal image set that could reliably create a high-quality model, reducing runtime and resource requirements. Dense panicle models could also be put to other uses - in addition to extracting counts, other phenotyping or health characteristics could be evaluated, perhaps based on crop volume, color, or texture. The model could also be used to plan physical interactions between robots and the modelled crop. This fits into our lab's larger goal of modelling plants for analysis and manipulation. 




\section*{ACKNOWLEDGMENTS}
We would like to thank Stephen Kresovich, Rick Boyles, and the Clemson team for the panicles. This work was partially supported by: ARPA-E TERRA DE-AR0001134, USDA NIFA 20216702135974, NSF Robust Int. 1956163.

\bibliographystyle{IEEEtran} 
\bibliography{mybib}
\end{document}